\documentclass[letterpaper, 10 pt]{IEEEtran}
\IEEEoverridecommandlockouts

\usepackage{amsmath,dsfont,amssymb}
\usepackage{amsthm}
\usepackage{graphicx}
\usepackage{amsfonts}
\usepackage{float}

\usepackage[colorlinks = true,
            linkcolor = blue,
            urlcolor  = blue,
            citecolor = blue,
            anchorcolor = blue]{hyperref}
\usepackage{algorithmicx}
\usepackage{algorithm}
\usepackage{algpseudocode}
\usepackage{lipsum}
\usepackage{subfig}
\usepackage{amssymb}
\usepackage{bbm}
\usepackage{float}
\usepackage{balance}
\usepackage{color}
\theoremstyle{plain}

\theoremstyle{definition}
\newtheorem{defn}{\textbf{Definition}}

\newtheorem{exmp}{\textbf{Example}}

\theoremstyle{remark}
\newtheorem{rem}{\textbf{Remark}}

\allowdisplaybreaks
\usepackage{algpseudocode}
\algdef{SE}[DOWHILE]{Do}{DoWhile}{\algorithmicdo}[1]{\algorithmicwhile\ #1}%

\usepackage{hybridsystem}

\usepackage[normalem]{ulem}
\DeclareMathOperator*{\argmax}{argmax}
\newcommand{\be}{b}
\newcommand{\Be}{\mathcal{B}_\mathcal{E}}
\newcommand{\G}{\mathcal{G}}
\begin{document}

\title{\LARGE \bf
Decentralized Task and Path Planning for Multi-Robot Systems
}
\author{Yuxiao Chen, Ugo Rosolia, and Aaron D. Ames
\thanks{The authors are with the Department of Mechanical and Civil Engineering, Caltech,
        Pasadena, CA, 91106, USA. Emails:
        {\tt\small \{chenyx, urosolia, ames\}@caltech.edu}}
}

\maketitle
\begin{abstract}
We consider a multi-robot system with a team of collaborative robots and multiple tasks that emerges over time. We propose a fully decentralized task and path planning (DTPP) framework consisting of a task allocation module and a localized path planning module. Each task is modeled as a Markov Decision Process (MDP) or a Mixed Observed Markov Decision Process (MOMDP) depending on whether full states or partial states are observable. The task allocation module then aims at maximizing the expected pure reward (reward minus cost) of the robotic team. We fuse the Markov model into a factor graph formulation so that the task allocation can be decentrally solved using the max-sum algorithm. Each robot agent follows the optimal policy synthesized for the Markov model and we propose a localized forward dynamic programming scheme that resolves conflicts between agents and avoids collisions. The proposed framework is demonstrated with high fidelity ROS simulations and experiments with multiple ground robots.
\end{abstract}
\section{Introduction}\label{sec:intro}
The planning and control of multi-robot systems is an important problem in robotics \cite{arai2002advances,yan2013survey}, and its applications can be seen in transportation, logistic robots in manufacturing and e-commerce, rescue missions post disasters, and multi-robot exploration tasks. The planning and control of the robotic agents is a core functionality of the multi-robot system, including the high-level task planning and the low-level path planning and control. Take the famous Kiva warehouse robot as an example \cite{wurman2008coordinating}, the task planning layer determines which robot shall pick up which package, then the path planning layer plans the specific trajectory for each robot in a grid, and the control module tracks the trajectory. Comparing to single robot operations, the core challenges of multi-robot systems are task allocation among multiple robot agents, and the trajectory planning that resolves conflicts between the robot agents.

The problem of task allocation for multi-robot systems has been studied extensively in the literature \cite{khamis2015multi,mataric2003multi}. Existing task allocation methods include the auction or market based methods \cite{tang2007complete,lagoudakis2004simple,bertsekas2009auction}, and optimization-based methods such as mixed-integer programming \cite{darrah2005multiple} and generic optimization algorithms \cite{mosteo2006simulated,liu2012centralized}. The drinking philosopher problem is utilized for coordination of multiple agents in \cite{sahin2020drinking}.

Another aspect of task allocation methods is whether they are centralized or decentralized. In the Kiva case, both the task assignment and the path planning are performed centrally, yet this may not be available if the multi-robot system operates in an environment without powerful sensing and computation capabilities. In general, the market/auction-based methods can be solved decentrally \cite{walsh1998market}, but the problem structure needs to be simple enough so that each agent can act as bidders and place their bids on the tasks, which may be difficult when some tasks require multiple agents to cooperate. The optimization-based methods allow for more complicated problem structures, yet may be difficult to solve in a decentralized manner. In \cite{choi2010decentralized}, the authors proposed the consensus-based bundle algorithm that allows for tasks requiring two agents to complete. However, it requires the agents to enumerate the possible bundles of task allocation and then resolve the conflict to achieve consensus, which does not scale as the number of agents grows. One powerful algorithm is the max-sum algorithm, which is based on the generalized distributive law (GDL) \cite{aji2000generalized}. Other instances of GDL algorithms include the max-product, the sum-product, and the min-sum, and they have been widely used in problems such as belief propagation and factor graph optimization. The distributed nature of max-sum allows it to be used in decentralized optimizations, including task allocation \cite{delle2012deploying,macarthur2011distributed}, and coordination \cite{farinelli2017advanced}. The proposed approach DTPP is based on the max-sum algorithm, and we shall show how max-sum is used to optimize the expected team reward in a decentralized manner.

\begin{figure}
  \centering
  \includegraphics[width=1\columnwidth]{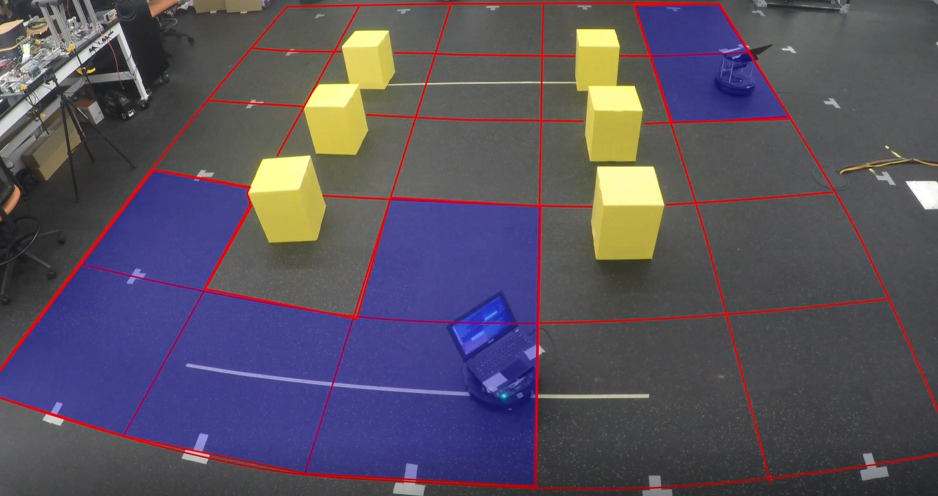}
  \caption{Experiment with a multi-robot system}\label{fig:expr}
  \vspace{-0.5cm}
\end{figure}

Compared to task allocation, the complexity of the motion planning problem for multi-robot system is typically higher. Depending on the representation of the state space, the problem can be either continuous or discrete. The continuous multiagent motion planning problem is typically nonconvex and therefore more emphasis is put on feasibility and safety rather than optimality. Several methods have been proposed to solve for collision-free trajectories via local resolution such as the velocity space methods \cite{ji2007computational,lumelsky1997decentralized}, and control barrier functions \cite{chen2020guaranteed}. The discrete multiagent motion planning problem deals with multiple agents on a graph, and the decisions are discrete \cite{wagner2011m}. Even in this case, it was shown in \cite{yu2013structure} that the multiagent motion planning problem on a graph is NP-hard.

The multi-robot system planning problem has been extended to the case with temporal logic specifications. STAP \cite{schillinger2018simultaneous} decomposes linear temporal logic (LTL) formulae into sub-tasks for a multi-robot system to perform simultaneous task allocation and planning. \cite{kloetzer2016multi} focused on syntactically co-safe LTL, and \cite{faruq2018simultaneous,nilsson2018toward} extend it to stochastic models such as Markov Decision Process (MDP) and Partially Observed Markov Decision Process (POMDP).

\begin{figure}
\vspace{-0.2cm}
  \centering
  \includegraphics[width=0.9\columnwidth]{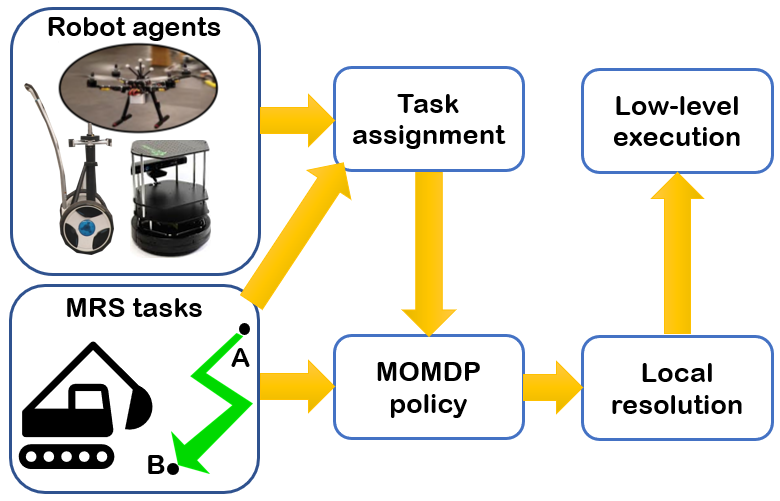}
  \caption{Overview of system structure}\label{fig:overview}
  \vspace{-0.2cm}
\end{figure}

\newsec{Contribution} We propose the DTPP framework that aims at maximizing the expected pure reward of a multi-robot system, where each robot's transition is modeled as an MDP or a MOMDP. The overall system structure is shown in Fig. \ref{fig:overview}. The task allocation module takes in the set of robot agents and multi-robot tasks and determines which task each robot agent commits to. Each robot agent then picks its action based on the MOMDP policy of the task it commits to, which may be modified by the local resolution module if there is a potential conflict with other robot agents. The high-level commands are then executed by the low-level controllers.

The contributions are threefold.
\begin{itemize}
  \item in Section \ref{sec:setup} we introduce the multi-robot tasks which are capable of describing tasks for multiple robot agents, potentially requiring coordination.
  \item Section \ref{sec:max_sum} presents the task allocation module that solves the task allocation with the highest expected pure reward for the multi-robot system, defined as the expected reward minus the expected cost. The algorithm is based on the max-sum algorithm, which is fully decentralized.
  \item Section \ref{sec:local_res} presents a local resolution module that locally resolves conflicts between agents when they are close to each other using a forward dynamic programming (DP) approach.
\end{itemize}
Simulation and experiment results are shown in Section \ref{sec:result}.

\section{Multi-robot system agents and tasks}\label{sec:setup}
We consider the problem of multi-robot operations consisting of multiple robot agents and multiple tasks that appear over time. The tasks are confined within an environment and can be accomplished within a finite time horizon, such as surveillance over a region, pickup and place, and collecting objects. Tasks that requires infinite time to complete, such as visit two points infinitely often, are not in the scope of this paper. We focus on the high-level task and path planning for the robotic agents, which contains two subproblems to solve: (1) task assignment of multiple tasks among the multiple robot agents (2) path planning for the robot agents.

\newsec{Robot agent modeling.}
The key idea is to extend the planning methods developed for a single robot to a multi-robot system. We use \textit{Mixed Observable Markov Decision Process} (MOMDP) to model a single robot planning problem, which is a tuple $\left( \mathcal{S}, \mathcal{E}, \mathcal{A},\mathcal{O}, T_s, T_e, O,J \right)$, where
\begin{itemize}

	\item $\mathcal{S}=\{1,\ldots,|\mathcal{S}|\}$ is a set of \textit{fully observable states};
	
	\item $\mathcal{E}=\{1,\ldots,|\mathcal{E}|\}$ is a set of \textit{partially observable states};

    \item $\mathcal{A}=\{1,\ldots,|\mathcal{A}|\}$ is a set of \textit{actions};

	\item $\mathcal{O}=\{1, \ldots,|\mathcal{O}|\}$ is the set of \textit{observations} for the partially observable state $e\in \mathcal{E}$;
			
	\item $T_s:\mathcal{S} \times \mathcal{A}\times \mathcal{E}\times \mathcal{S}\rightarrow [0,1]$ is the \textit{observable state transition probability function} where $T_s(s,a,e,s')$ is the probability of the transition from $s$ to $s'$ under action $a$ and partially observable state $e$.

    \item $T_e: \mathcal{S} \times\mathcal{E}\times \mathcal{A}\rightarrow [0,1]$ is the \textit{partially observable state transition probability function} where $T_e(s,e,a,e')$ is the probability of the transition from $e$ to $e'$ given the action $a$ and the current observable state $s$

	\item $O:\mathcal{S}\times \mathcal{E} \times \mathcal{A} \times \mathcal{O} \rightarrow [0,1]$ is the \textit{observation function} where $O(s,e,a,o)$ describes the probability of observing the measurement $o\in \mathcal{O}$, given the current state of the system $(s,e)$ and the action $a$ applied at the previous time step
    \item $J:\mathcal{S}\times\mathcal{A}\times\mathcal{S}\times\mathcal{O}\to\mathbb{R}$ is the \textit{cost function} where $J(s,a,s',o)$ is the cost associated with the transition from $s$ to $s'$ under action $a$ with observation $o$.

\end{itemize}

\begin{figure}
  \centering
  \includegraphics[width=1\columnwidth]{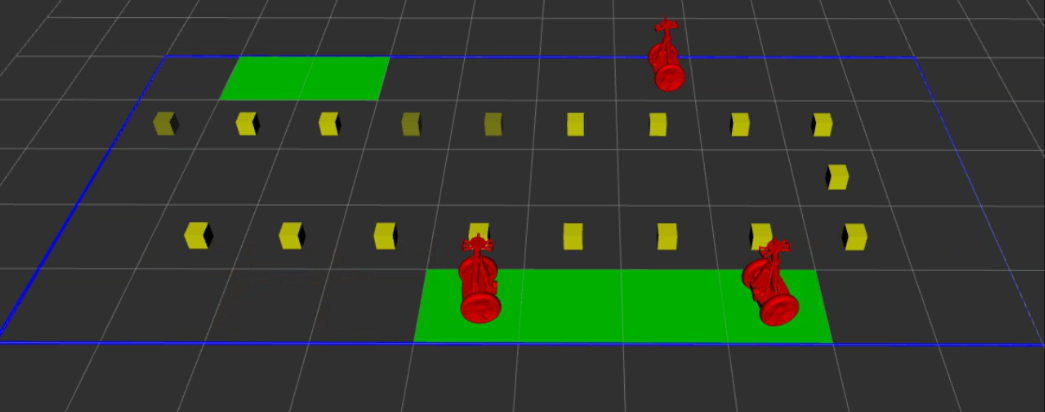}
  \caption{ROS simulation of 3 segway robots with 3 uncertain regions}\label{fig:sim}
  \vspace{-0.5cm}
\end{figure}

It is assumed that there exists an idle action, denoted as $\text{IDLE}\in\mathcal{A}$, which keeps the robot agent at the current state and incurs no cost.
\begin{rem}
  Note that the MOMDP is only given a cost function, this is because all the robot agents incur the same cost as they move in the environment, but the rewards for accomplishing the tasks differ in different tasks, and the same MOMDP is used to describe all tasks.
\end{rem}
\begin{rem}
  When all states are observable, the MOMDP is reduced to a Markov Decision Process (MDP).
\end{rem}
The MOMDP is shared by all agents in the multi-robot system where each robot agent selects an action at each time step, and the actions are executed simultaneously.  A multi-robot system is then abstracted as a tuple $(\mathcal{I},\text{MOMDP})$, where $\mathcal{I}=[1,2,...,N]$ is the indices of the robot agents. The overall cost for the multi-robot system is then the summation of the individual costs based on $J$. A collision happens when two robot agents are at the same state the same time, and we shall use the local resolution scheme presented in Section \ref{sec:local_res} to prevent collisions.

In the example used in this paper, the observed state is the position of the robot agent, the partially observable state is whether some regions of the environment are blocked with obstacles or free to pass, such as the half transparent obstacles in Fig. \ref{fig:sim}. The robot agent can get stochastic observations of the uncertain state, which gets more deterministic as the robot gets closer to the region.

\newsec{Multi-robot tasks.}
A multi-robot task is a tuple $(\mathcal{G},\mathcal{J},t_0,t_f,\mathcal{R})$, where $\mathcal{G}$ is the goal set of the task, $\mathcal{J}\subseteq\mathcal{I}$ is the set of robot agents involved in this task, which is understood as the candidates for completing the task. $t_0$ is the starting time of the task and $t_f$ is the ending time of the task. By $t_f$, the reward shall be collected based on the arrival of robot agents to the goal set of the task. The reward function $\mathcal{R}:\{0,1\}^{|\mathcal{J}|}\to\mathbb{R}$ maps the arrival status of the robot agents to reward value. In the homogeneous case, $\mathcal{R}$ is simply a function of the number of robot agents arriving at the goal set before $t_f$, while in the heterogeneous case, different robot agents can incur different reward. For simplicity, we focus on the homogeneous case for the remainder of this paper. Let $(r_0,r_1,r_2...)$ be the compact form of $\mathcal{R}$, where $r_i$ denotes the reward with $i$ agents arriving. Albeit simple, the reward function $\mathcal{R}$ can be quite expressive, here are a few examples.

\begin{exmp} $(r_0=0,r_1=5,r_2=5)$ can represent a surveillance task, one robot arriving at the goal set is sufficient for the task, additional robots arriving would not incur additional reward.
\end{exmp}

\begin{exmp} $(r_0=0,r_1=5,r_2=8)$ can represent the task of moving a pile of sand with the total weight of 8 kg, yet one robot can only carry 5 kg. Therefore, $r_1=5$ and $r_2=8$.
\end{exmp}

\begin{exmp} $(r_0=0,r_1=0,r_2=8)$ can represent the task of moving a box that weighs 8 kg, yet one robot can only carry 5 kg. Therefore, one robot arriving cannot move the box, which incurs zero reward, and two robots arriving shall collect the full reward.
\end{exmp}
\noindent We shall show in Section \ref{sec:max_sum} how the reward function is combined with the MOMDP to maximize the expected pure reward.

\section{Dynamic task allocation with max-sum}\label{sec:max_sum}
\newsec{The pure reward of a multi-robot system.}
Given the MOMDP that describes the robot transition dynamics and a multi-robot task, ideally, one would construct the product MDP/MOMDP for the whole multi-robot system and plan the joint action, yet this is usually not implementable due to the doubly exponential complexity~\cite{belta2017formal}. Instead, we use one single MDP/MOMDP for a task assuming the agents' evolution is independent of each other and let all agents committed to the task run the same policy in parallel. Obviously, the assumption is not true in practice as we use the local resolution scheme to prevent collisions between agents. However, when the multi-robot system is scattered with a relatively low density, i.e., the interactions between agents are not frequent, this assumption can be quite close to reality.
\begin{rem}
  To reflect the potential influence of the local resolution, a higher probability of staying at the current state is assigned to the MOMDP introduced in Section \ref{sec:setup}.
\end{rem}

\newsec{MOMDP policy synthesis.} We use an optimal quantitative approach to synthesize the policy for the MOMDP where the policy optimizes the cost function over all policies that maximize the probability of satisfying the specification, which is to reach the goal set before the terminal time. Given a MOMDP, a goal set $\mathcal{G}$, and a horizon $t_f$ ($t_0$ is set to 0 for notational simplicity), the optimal quantitative synthesis problem is the following:
\begin{equation}\label{eq:MOMDP_opt}
\begin{aligned}
\pi^{\star} = \mathop{\arg\max}\limits_{\pi} ~~ & \mathbb{E}^\pi \Bigg[ \sum_{t = 0}^{t_f-1} -J(s_{t},a_{t},s_{t+1},o_{t+1}) \Bigg]   \\
 \text{subject to} ~~ ~& \pi \in {\arg\max}_{\kappa} \mathbb{P}^\kappa\Bigg[\bigvee\limits_{t=0}^{t_f}s_t\in\mathcal{G}\Bigg],
\end{aligned}
\end{equation}
which solves for the policy that minimizes the expected cost among all policies that maximize the probability of reaching the goal set $\mathcal{G}$ before $t_f$. Problem~\eqref{eq:MOMDP_opt} is using a point-based strategy as in~\cite{isom2008piecewise,pineau2003point}, see Appendix \ref{sec:quant} for more detail. $\pi:\mathcal{S}\times\mathcal{B}_{\mathcal{E}}\to \mathcal{A}$ is a policy for the MOMDP that maps the current state and the belief vector $b$ to an action, where $b\in\mathcal{B}_{\mathcal{E}}\doteq\{b\in\mathbb{R}^{|\mathcal{E}|}_{\ge 0}\mid \mathds{1}^\intercal b=1\}$.

The solution of \eqref{eq:MOMDP_opt} consists of two value functions $V_J$ and $V_\mathcal{G}$, and the optimal policy $\pi^\star$. The two value functions have clear physical meanings:
\begin{equation}\label{eq:value}
  \begin{aligned}
  V_J(t,s,b)&=\mathbb{E}^{\pi^\star}[\sum_{\tau=t}^{t_f}-J(s_{\tau},a_{\tau},s_{\tau+1},o_{\tau+1})|e_t\sim b]\\
  V_\mathcal{G}(t,s,b)&=\mathbb{P}^{\pi^\star}[\bigvee\limits_{\tau=t}^{t_f}s_\tau\in\mathcal{G}|e_t\sim b],
  \end{aligned}
\end{equation}
that is, given the time, state, and belief vector, $V_J$ represents the expected negative cost-to-go, and $V_\mathcal{G}$ represents the probability of reaching the goal set before $t_f$. in the MDP case, $V_J$ and $V_\mathcal{G}$ would be functions of only $t$ and $s$.

With these two value functions, the expected pure reward of a task can then be approximated given the robot agents committed to the task. In the homogeneous agent case, this is simply
\begin{equation}\label{eq:ER}
\mathbb{E}[R]=\sum\nolimits_{i=0}^{|\mathcal{J}|} r_i P_c[i] - \sum\nolimits_{j\in\mathcal{J}} V_J(t,s^j_t,b),
\end{equation}
where $\mathcal{J}\subseteq\mathcal{I}$ is the set of robot agents committed to the task, $P_c[i]$ is the cumulative probability of exactly $i$ agents arriving at the goal set by $t_f$, and is calculated as
\begin{equation*}
p_i=\sum\nolimits_{c^j\in\mathbb{B},j\in\mathcal{J}|,\sum\limits_j c^j=i} V_\mathcal{G}(t,s^j_t,b)^{c^j}(1-V_\mathcal{G}(t,s^j_t,b))^{1-c^j}.
\end{equation*}
For example, suppose $\mathcal{J}=\{1,2\}$ and $p_j=V_\mathcal{G}(t,s^j_t,b),j=1,2$ are the probabilities of the two agents reaching the goal set by $t_f$ from their current state and time, respectively, which are directly obtained from $V_\mathcal{G}$. Then $P_c$ is calculated as
\begin{equation*}
\begin{aligned}
  P_c[0]&=(1-p_1)(1-p_2)\\
  P_c[1]&=p_1(1-p_2)+(1-p_1)p_2,\\
  P_c[2]&=p_1 p_2.
  \end{aligned}
\end{equation*}

Given a multi-robot system with a set $\mathcal{K}$ of multiple tasks, let $M^i(t)$ be the set of all tasks that involve robot agent $i$ at time $t$ plus $\emptyset$, $m^i\in M^i(t)$ be the commitment variable where $m^i=k$ indicates that robot agent $i$ is committed to task $k$, and $m^i=\emptyset$ indicates that robot agent $i$ is not committed to any task. When $m^i=\emptyset$, it is assumed that the robot agent would stay still and incurs zero cost.  The expected pure reward for each task then can be computed with \eqref{eq:ER}. We let $F_k(\{m^i\}_{\mathcal{J}_k})$ denote the expected pure reward of task $k$ as a function of the commitment of the robot agents in the candidate set $\mathcal{J}_k$. Note that each $\mathcal{J}_k\subseteq \mathcal{I}$ and they may have overlaps, i.e., one agent can be included in the candidate set of multiple tasks. The simplest choice is to take $\mathcal{J}_k=\mathcal{I}$ for all $k\in\mathcal{K}$, but in practice, one can exclude some robot agents with little chance of completing the task (e.g. agents that are too far away), which accelerates the computation.

\newsec{Factor graph and the max-sum algorithm.} It can be easily verified that the total expected pure reward is $\sum_{k\in\mathcal{K}} F_k(\{m^i\}_{\mathcal{J}_k})$, which is a function of $\{m^i\}_{\mathcal{I}}$. Note that the expected reward for each task is calculated based on the commitment of the robot agents, and the expected cost of each agent is summed up except the ones that are not committed to any tasks, which incurs zero cost. The dynamic task assignment problem is then to determine the commitment of the robot agents that leads to the largest expected pure reward:
\begin{equation}\label{eq:task_assignment}
  \mathop{\max}\limits_{\{m^i\}_{\mathcal{I}}} \sum_{k\in\mathcal{K}}F_k(\{m^i\}_{\mathcal{J}_k}).
\end{equation}
The task assignment is ``dynamic'' because \eqref{eq:task_assignment} changes over time, and is solved in every time step.

\begin{figure}
  \centering
  \includegraphics[width=0.6\columnwidth]{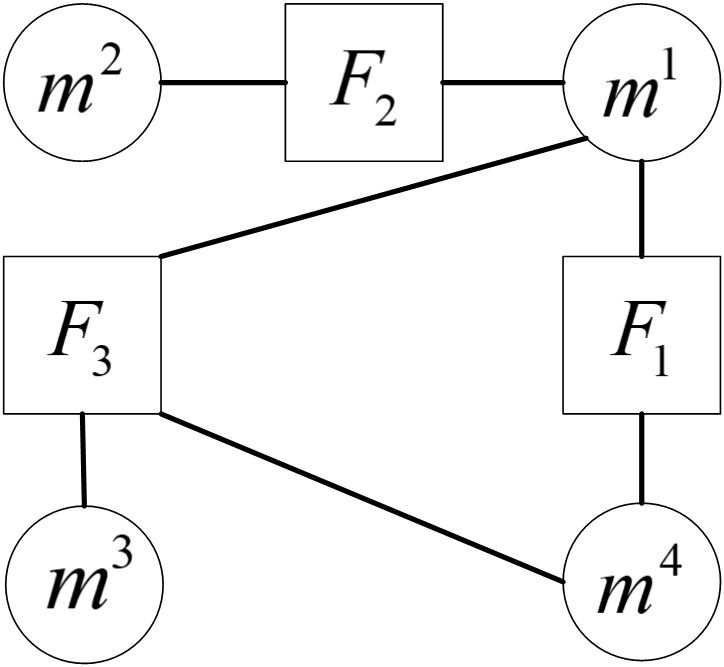}
  \caption{Factor graph of 4 robot agents and 3 tasks}\label{fig:factor_graph}
\end{figure}

To this point, \eqref{eq:task_assignment} is in the form of a factor graph, which is a bipartite graph representing the factorization of a function. It contains two types of nodes, variable nodes and factor nodes. In our case, the variable nodes are the commitment $m^i$ of the robot agents, and the factor nodes are the expected pure reward $F_k$ of each tasks. As an example, Fig. \ref{fig:factor_graph} shows the factor graph with 4 robot agents and 3 multi-robot tasks, where $\mathcal{J}_1=\{1,4\},~\mathcal{J}_2=\{1,2\},~\mathcal{J}_3=\{1,3,4\}$.
We then use the max-sum algorithm to solve the task assignment problem, similar to \cite{macarthur2011distributed}. The max-sum algorithm seeks to maximize the sum of all factors via exchanging messages between the factor nodes and the variable nodes. To be specific, two types of messages are exchanged: the $q$ messages from variables to factors, and the $r$ messages from factors to variables.
\begin{equation}\label{eq:max_sum_msg}
\begin{aligned}
q_{i\to k}(m^i)&=\alpha_{ik}+\sum_{n\in M^i\backslash k} r_{n\to i}\\
r_{k\to i}(m^i)&=\mathop{\max}\limits_{\mathcal{J}_k\backslash i}[F_k(\{m^i\}_{\mathcal{J}_k})+\sum_{n\in \mathcal{J}_k\backslash i} q_{n\to k}(m^n)].
\end{aligned}
\end{equation}
All the messages are exchanged locally and no central coordination is needed. Once the messages converge, the optimal solution can be solved as
\begin{equation}\label{eq:max_sum_opt}
  {m^i}^\star = \arg\max \sum_{k\in M^i} r_{k\to i}(m^i).
\end{equation}

\begin{algorithm}
    \caption{Task Allocation with max-sum}
    \label{alg:max_sum}
    \begin{algorithmic}[1] 
        \Procedure{Task\_allocation}{$\{s^i_t\}_{\mathcal{I}}$, $\mathcal{K}$, $b_t$, Maxiter}
            \For{$k\in\mathcal{K}$}
                \For{$\{m^i\}_{\mathcal{J}^k}$ in $\{M^i\}_{\mathcal{J}^k}$}
                    \State Calculate value table entry $F_k(\{m^i\}_{\mathcal{J}^k})$
                \EndFor
            \EndFor
            \State  iter$\gets 0$
            \While{iter$<$Maxiter}
                \For{$i\in\mathcal{I}$}
                    \State Update $q$ messages with \eqref{eq:max_sum_msg}
                \EndFor
                \For{$k\in\mathcal{K}$}
                    \State Update $r$ messages with \eqref{eq:max_sum_msg}
                \EndFor
                \If{$r$ and $q$ messages do not change}
                    \State \textbf{Break}
                \EndIf
                \State iter++
            \EndWhile
            \For{$i\in\mathcal{I}$}
                \State Calculate ${m^i}^\star$ with \eqref{eq:max_sum_opt}
            \EndFor
            \State \Return $\{m^i\}_{\mathcal{I}}$
        \EndProcedure
    \end{algorithmic}
\end{algorithm}

The max-sum algorithm is guaranteed to converge if the factor graph is acyclic. Although there is no convergence guarantee on cyclic graphs, multiple empirical studies show that the solution quality is decent even when it does not converge to the optimal solution. Moreover, there exist variations of max-sum that return suboptimal solutions with a bounded optimality gap~\cite{rogers2011bounded}.

\section{Local path planning}\label{sec:local_res}
One key assumption we made is that the evolution of the robot agents is independent of each other, which decompose the multi-robot system planning problem into multiple single-agent planning problem. As pointed out in previous sections, this is not true in practice due to the collision avoidance constraint. We shall present a local resolution scheme to coordinate adjacent robot agents and avoid a collision. The first step is to construct the adjacency graph for the multi-robot system.
\begin{defn}
  Given an MOMDP, two states $s,s'$ are adjacent if $\exists a,a'\in\mathcal{A}, e\in\mathcal{E},s''\in\mathcal{S}$ such that $T_s(s,a,e,s'')>0,$ $T_s(s',a',e,s'')>0$, that is, two states are adjacent if there exist actions for the two states under which their possible successor states intersect.
\end{defn}

\newsec{Forward dynamic programming for local conflict resolution.} Given the multi-robot system, two agents $i,j\in\mathcal{I}$ are adjacent if their current state $s^i_t,s^j_t$ are adjacent. Let $G$ be the adjacency graph with the nodes being the robot agents $\mathcal{I}$, and two nodes are connected if they are adjacent. $G$ is divided into connected subgraphs, and for each connected subgraph, if it only contains one node, the robot agent simply follow the policy of the task it committed to; if it contains more than one node, the local resolution scheme is used to resolve the conflict.

Note that any subgraph only needs to consider the nodes within the subgraph since, by construction, the nodes will not collide with nodes outside the subgraph. Let $\bar{\mathcal{I}}$ be the robot agents within one connected subgraph, the local resolution problem at time $t$ is the following:
\begin{equation}\label{eq:local_res}
\resizebox{1\hsize}{!}{$
\begin{aligned}
  \mathop{\max}\limits_{\{a^i_{t:t+T}\}_{\bar{\mathcal{I}}}}&\sum_{i\in\bar{\mathcal{I}}}\mathbb{E} \left[\begin{gathered} \sum_{\tau=t}^{t+T-1}-J(s^i_\tau,a^i_\tau,s^i_{\tau+1},o^i_{\tau+1})+\\ V^i_J(t+T,s^i_{t+T},b_t)+\delta R^i V^i_\mathcal{G}(t+T,s^i_{t+T},b_t)\end{gathered}\right]\\
  \mathrm{s.t.}~ & \forall i\in \bar{\mathcal{I}},t\le\tau\le t+T-1,s^i_{\tau+1}\sim\sum_e T_s(s^i_\tau,a^i_\tau,e)b_t(e)\\
  & \forall i,j\in \bar{\mathcal{I}},i\neq j,\forall \tau\in\{t,...,t+T\},\mathbb{P}(s^i_\tau=s^j_\tau)=0,
\end{aligned}
$}
\end{equation}
where $T$ is the look-ahead horizon of the forward dynamic programming (DP), $b$ is the belief vector at time $t$. Since we don't have access to future observations, $b_t$ is assumed constant over the horizon. $\delta R^i$ is the discrete derivative of the task reward that agent $i$ is committed to, i.e., the reward difference agent $i$ would make if it arrives at the goal set, which can be computed given the reward function $\mathcal{R}$ of the task. $V^i_J$ and $V_\mathcal{G}^i$ are the two value functions associated with the task that agent $i$ commits to. \eqref{eq:local_res} is a sequential decision making problem with running reward $-J$ and terminal reward $\sum_{i\in\bar{\mathcal{I}}}R^i V^i_\mathcal{G}+V^i_J$, which is the expected reward at the terminal state.
\begin{algorithm}
    \caption{Forward DP for local resolution}
    \label{alg:forward_DP}
    \begin{algorithmic}[1] 
        \Procedure{Loc\_res}{$\{s^i_t,\delta R^i,V^i_J,V^i_{\mathcal{G}}\}_{\bar{\mathcal{I}}}$,\ $b_t$}
            \State Initialize the search tree $\mathcal{T}$ with $\{s^i_t\}_{\bar{\mathcal{I}}}$
            \For{$\tau=t,...t+T-1$}
                \State Expand $\mathcal{T}$ with all action combinations
                \State Trim collision nodes and dominated nodes from $\mathcal{T}$
            \EndFor
            \State Add terminal reward to the leaf nodes
            \State \Return $\{a^i_{t}\}_{\bar{\mathcal{I}}}$ associated with the optimal leaf node
        \EndProcedure
    \end{algorithmic}
\end{algorithm}

\newsec{Algorithms.} The forward DP algorithm is summarized in Algorithm \ref{alg:forward_DP}, where the search tree consists of nodes that store the collective state distribution of all agents in $\bar{\mathcal{I}}$ and the current cumulated reward and edges that store the joint actions. The trimming procedure removes nodes that contain possible collisions and nodes whose cumulated reward is smaller than another node sharing the same state distribution. Compared to backward DP, since the DP horizon, $T$ is typically chosen to be small, forward DP saves computation time because not all states in the state space are explored. Algorithm \ref{alg:forward_DP} runs in a receding horizon fashion, i.e., only the first step of the action sequence is executed, and the algorithm replans in every time step.

To implement Algorithm \ref{alg:forward_DP} in a decentralized setting, one can simply select a node within the subgraph as the host and perform Algorithm \ref{alg:forward_DP} and share the result with other nodes in the subgraph.

Algorithm \ref{alg:MRS_planning} summarizes all the modules of the DTPP, where $\mathcal{I}_j[1]$ is the only element in $\mathcal{I}_j$ when $|\mathcal{I}_j|=1$. Besides the procedures introduced in Algorithm \ref{alg:max_sum} and \ref{alg:forward_DP}, other procedures involved are
\begin{itemize}
  \item \textsc{Obtain\_partition} takes the current state of all agents and calculates the adjacency graph, then returns node sets in all connected subgraphs, denoted as $\{\mathcal{I}_j\}$
  \item \textsc{Policy} evaluates the optimal policy of the task that the agent commits to
  \item \textsc{Execute} executes the action and obtain the next state and observation
  \item \textsc{Update\_belief} updates the belief with the new state and observation obtained from executing the action
  \item \textsc{Update\_task} updates the task set, removing expired tasks and adding new tasks should there be any.
\end{itemize}

Note that the belief gets updated sequentially by all the agents after executing their actions, and this piece of information is shared among the whole multi-robot system.

\begin{algorithm}
    \caption{multi-robot system planning}
    \label{alg:MRS_planning}
    \begin{algorithmic}[1] 
    \State \textbf{Input:} $(\mathcal{I},\text{MOMDP})$ $\mathcal{K}_0$, $b_0$, $\{s_0^i\}_{\mathcal{I}}$, Maxiter
    \State $t\gets 0$, $\mathcal{K}\gets \mathcal{K}_0$
    \While{Not Terminate}
        \State $\{m^i\}_{\mathcal{I}}$=\textsc{Task\_allocation}{$\{s^i_t\}_{\mathcal{I}}$, $\mathcal{K}$, $b_t$, Maxiter}
        \State $\{\mathcal{I}_j\}$=\textsc{Obtain\_partition}($\{s_t^i\}_{\mathcal{I}}$, MOMDP)
        \For{$\mathcal{I}_j\in \{\mathcal{I}_j\}$}
            \If{$|\mathcal{I}_j|==1$}
                \State $i\gets\mathcal{I}_j[1]$ the robot agent index in
                \If{$m^i==\emptyset$}
                    \State $a_{t}^i\gets\text{IDLE}$
                \Else
                    \State $a_{t}^i\gets$ \textsc{Policy}($t,m^i,s^i_t,b_t$)
                \EndIf
            \Else
                \For{$i\in\mathcal{I}_j$}
                    \State Obtain $\delta R^i,V^i_J,V^i_{\mathcal{G}}$ from task set $\mathcal{K}$
                \EndFor
                \State $\{a^i_{t}\}_{\mathcal{I}_j}\gets$ \textsc{Loc\_res}($\{s^i_t,\delta R^i,V^i_J,V^i_{\mathcal{G}}\}_{\mathcal{I}_j}$,\ $b_t$)
            \EndIf
        \EndFor
        \For{$i\in\mathcal{I}$}
            \State $o^i_{t+1},s^i_{t+1}\gets$ \textsc{Execute}($i,a_t^i$)
            \State $b_{t}\gets$ \textsc{Update\_belief}($s_t^i,a_t^i,s_{t+1}^i,o_{t+1}^i,b_{t}$)
        \EndFor
        \State $b_{t+1}\gets b_t$, $t\gets t+1$
        \State $\mathcal{K}\gets$ \textsc{Update\_task}($\mathcal{K},t$)
    \EndWhile
    \end{algorithmic}
\end{algorithm}

\section{Results}\label{sec:result}
We demonstrate the proposed DTPP framework with a grid world example both in simulation and in experiments.

\newsec{Multi-robot system setup.}
   The MOMDP is setup as a grid world. Each robot agent can choose from 5 actions: $\mathcal{A}=\{N,S,W,E,\text{IDLE}\}$, which make the robot move north, south, west, east, and stay still. The observable state is the agents' position within the grid world, and the partially observable state is the obstacle status of several uncertain cells, which may be clear or occupied by an obstacle. $\mathcal{E}=\mathbb{B}^{N_u}$, where $N_u$ is the number of uncertain cells, and $|\mathcal{E}|=2^{N_u}$. The state transition is assumed to be deterministic, i.e., for any state $s$ and action $a$, there is only one possible successor state.
\begin{rem}
  The state transition is deterministic when executing the actions, however, the agent is assumed to have 10\% chance of staying still with $a\in\{N,S,W,E\}$ when solving for the quantitative optimal policy. This is to account for the possible influence of the local resolution and is particularly important for $V_\mathcal{G}$, the probability of reaching the goal set. Without the change of probability, the agent might think that it has 100\% chance of reaching the goal yet fail to do so due to the local resolution preventing it from executing the action according to the policy. Under this change, the agent will be more certain that it can reach the goal as it gets closer to the goal.
\end{rem}
  The observation $\mathcal{O}$ space is the same as $\mathcal{E}$, and for each of the uncertain cells, we have
\begin{equation*}
  \forall i\in\{1,...,N_u\},\mathbb{P}(o_i=e_i)=\left\{ {\begin{array}{*{20}{c}}
  {1}&{d\le 1} \\
  {0.8}&{d=2} \\
  {0.5}&{d>2}
\end{array}} \right.,
\end{equation*}
where $o_i$ and $e_i$ are the $i$th entry of $o$ and $e$, the observed state and actual state of the uncertain cells, $d$ is the Manhattan distance from $s$ to the $i$th uncertain cell. $T_e$ is set so that each uncertain cell has a 0.05 chance of changing its current state (from obstacle to free or the other way) when no agents are 2 steps or closer to it. The cost function $J$ gives penalty 1 to all actions but IDLE.

\begin{figure}
  \centering
  \includegraphics[width=1\columnwidth]{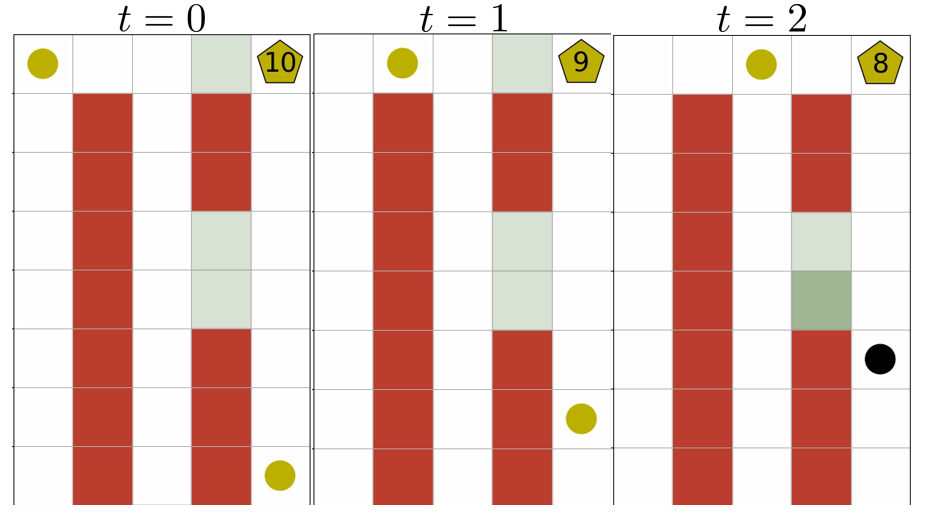}
  \caption{Simulation with two agents and one task}\label{fig:2agents}
\end{figure}

Fig. \ref{fig:2agents} shows a sample simulation on a $7\times5$ grid world. The red blocks are the known obstacles and the green blocks are the uncertain cells with the transparency equal to the belief of it being an obstacle. The two yellow circles are the agents and the yellow pentagon denotes the goal region of the task, with the number showing the time left before $t_f$. The task's reward function $\mathcal{R}$ has a compact form of $(0,50,50)$, which means that one agent reaching shall earn a reward of 50, and two agents reaching will not increase the reward. The evolution of the value functions is shown in Table \ref{tab:example}.
\begin{table}[]
\begingroup
\setlength{\tabcolsep}{10pt} 
\renewcommand{\arraystretch}{1.5} 
\caption{Evolution of the value functions in the example shown in Fig. \ref{fig:2agents}}
\label{tab:example}
\begin{tabular}{cccc}
\hline
 &  $t=0$  &  $t=1$ &  $t=2$   \\ \hline
$\bigvee\nolimits_{t}^{t_f}s^1_t\in\mathcal{G}$ &  0.727  &  0.800 &0.999  \\
 $\bigvee\nolimits_{t}^{t_f}s^2_t\in\mathcal{G}$                       &  0.947  &  0.962 & 0.974  \\
$\bigvee\nolimits_{t}^{t_f}s^1_t\in\mathcal{G}\vee s^2_t\in\mathcal{G}$ &  0.985  &0.992 & 0.99997 \\
$V_J^1$                                                                   & -3.636 & -2.89 &  -2.22   \\
$V_J^2$                                                                  & -7.616 & -6.56 &  -5.49   \\ \hline
\end{tabular}
\endgroup
\end{table}

At $t=0$, both agents are assigned to the task because neither of them has 100\% chance of reaching the goal in time. The algorithm decides to put two agents on the task to increase the probability that at least one of them reaches the goal, which leads to a higher expected pure reward. At $t=2$, the agent on the top figured out that the uncertain cell blocking its path to the goal is clear, which significantly increase the probability of reaching the goal from 80\% to 99.9\%, then the max-sum algorithm decided that the lower agent can stay idle to safe the cost.

To demonstrate the applicability of DTPP on real robotic systems, we ran high-fidelity simulations with multiple Segway robots and performed experiments with turtlebots.

\newsec{Segway simulation.} In the Segway simulation, each Segway follows a nonlinear model with 7 states: $x=[X,Y,\theta,\dot{\theta},v,\psi,\dot{\psi}]^\intercal$, where $X,Y$ are the longitudinal and lateral coordinates, $\theta$ and $\dot{\theta}$ are the yaw angle and yaw rate, $\psi$ and $\dot{\psi}$ are the pitch angle and pitch rate, and $v$ is the forward velocity. The input vector consists of the torques on the two wheels.

The high-level planning follows Algorithm \ref{alg:MRS_planning} which sends high-level commands to the low-level controller, which runs a Model Predictive Controller (MPC) that generates torque command to the Segway. The high-level command consists of three parts: the desired waypoint $x^\star$, the state constraint $\mathcal{C}$, and the terminal state constraint $\mathcal{C}_f$. In the grid world case, $\mathcal{C}$ is simply the union of the current grid box and the next grid box to transition to, and $\mathcal{C}_f$ is the next grid box. The MPC then solves the following optimization to obtain the torque input:
\begin{equation}\label{eq:MPC}
  \begin{aligned}
  \mathop{\min}\limits_{u_{t:t+T-1}} &\sum_{\tau=t}^{t+T-1} x_\tau^\intercal Q x_\tau + u_\tau^\intercal R u_\tau + x_{t+T}^\intercal Q_f x_{t+T}\\
  \mathrm{s.t.} & \forall \tau=t,...t+T-1, x_{\tau+1} =f(x_\tau,u_\tau), \\
  &\forall \tau=t,...t+T-1, x_\tau \in\mathcal{C}, u_\tau\in\mathcal{U}, x_{t+T-1}\in\mathcal{C}_f,
  \end{aligned}
\end{equation}
where $T$ is the horizon of the MPC, $\mathcal{U}$ is the set of available input, and $f$ is the robot dynamics. The MPC uses sequential quadratic programming to accelerate the computation so that it can be implemented in real-time.

\begin{figure}
    \centering
    \includegraphics[width=1\columnwidth]{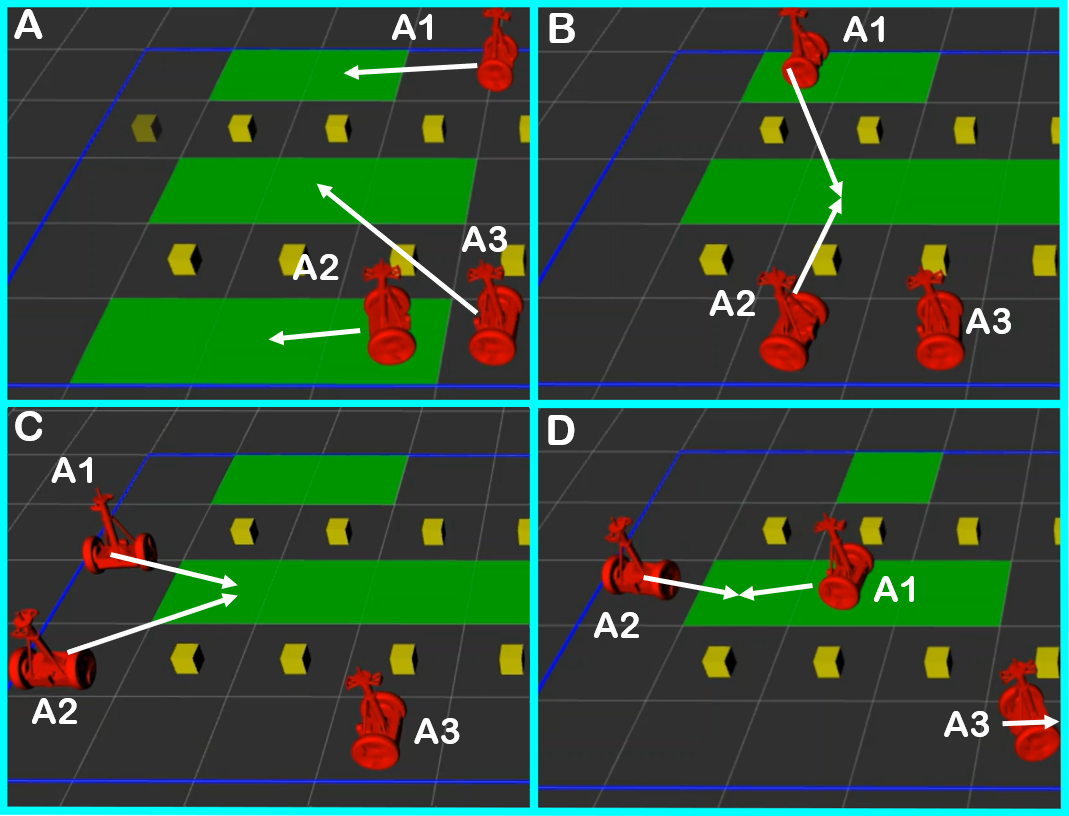}
    \caption{Simulation with three segway robots}\label{fig:sim_example}
    \label{fig:my_label}
\end{figure}

Fig. \ref{fig:sim_example} shows one scenario of the simulation. At the beginning, Agent 2 was assigned to an existing task while the new task with the goal region in the middle appears, and agent 3 committed to the new task. Then in the second frame, the task that agent 2 committed to expires, and DTPP decided to let agent 2 commit to the new task, and agent 3 turned idle. Then in the third frame, the local resolution module made sure that agent 1 and agent 2 do not collide and let agent 1 enter the goal region first. In the last frame, agent 1 did not stop after entering the goal region, but kept moving to make room for agent 2, and agent 3 committed to a new task out of the frame.

\newsec{Turtlebot experiment.} We conducted experiments using two turtlebots with differential driving capabilities. Since the model is simply a Dubin's car model with velocity and yaw rate inputs, we use a simple PID controller as the low-level controller for the turtlebots. The experiment is performed on a $5\times5$ grid with 3 uncertain cells, of which one is a free box, and the other two are filled with obstacles. The two turtlebots are presented with randomly generated multi-robot tasks with random goal regions and horizons ranging from 5 to 9 time steps. The high-level time step would increase by 1 if all agents reach the desired grid box planned by the high-level planner.

\begin{figure}
  \centering
  \includegraphics[width=1\columnwidth]{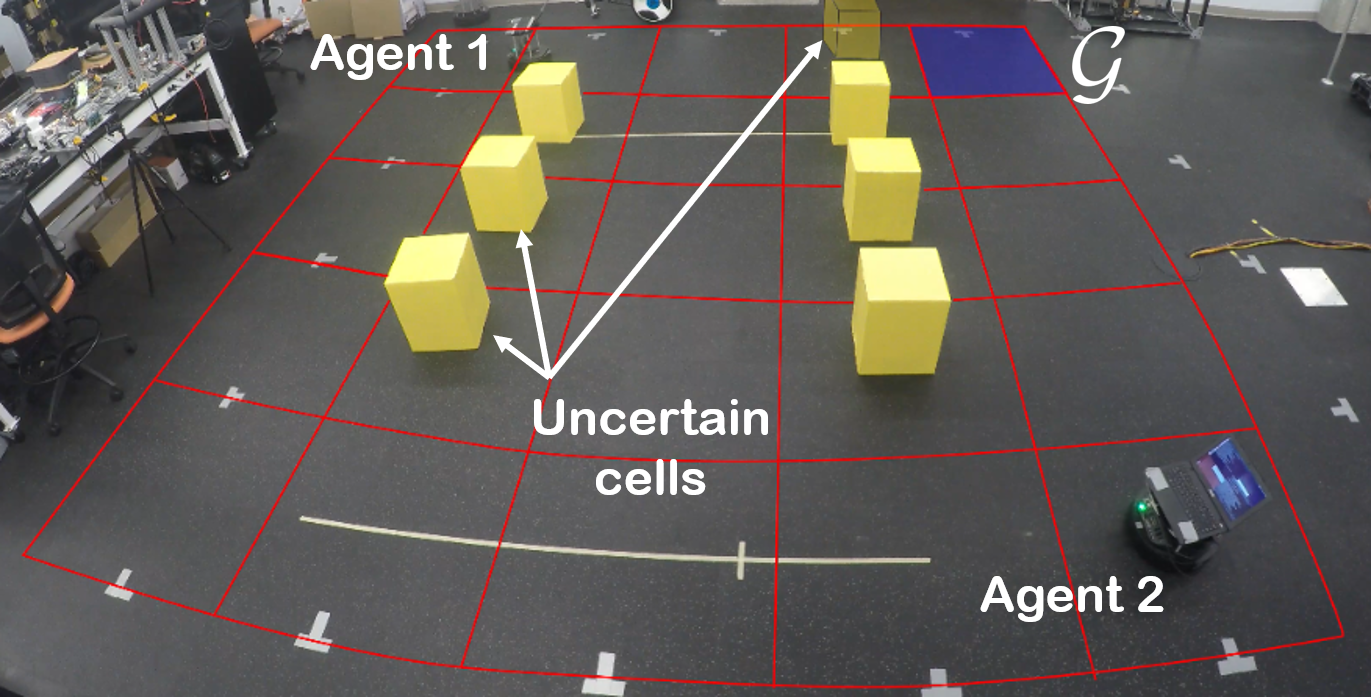}
  \caption{Experiment with 2 turtlebots in a $5\times 5$ grid world}\label{fig:experiment}
\end{figure}

Fig. \ref{fig:experiment} shows a situation similar to that described by Fig. \ref{fig:2agents}, where the robot agents figured out that the uncertain region is free of obstacle and the max-sum algorithm decides to let one robot agent go IDLE to save cost. A video containing the experiment and simulation result can be found in \href{https://youtu.be/zzvD-ukcsis}{link}.

\begin{figure}
  \centering
  \includegraphics[width=1\columnwidth]{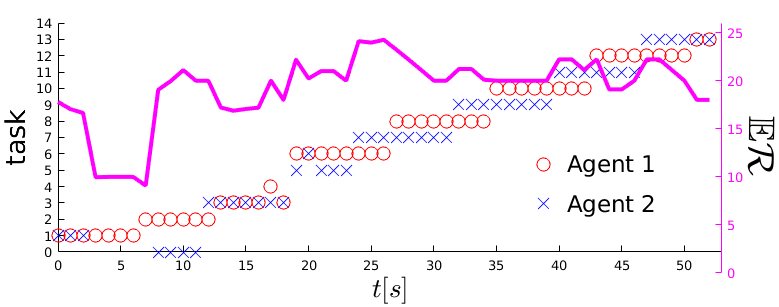}
  \caption{Task assignment and expected pure reward in the turtlebot experiment}\label{fig:exp_res}
\end{figure}

Fig. \ref{fig:exp_res} shows the task assignment result given by the max-sum algorithm and the expected pure reward of the multi-robot system consisting of two turtlebots during the experiment. We kept the number of tasks at every time instance to be 2, and the task reward for most tasks (randomly generated except the first 4 tasks) is $(r_0=0,r_1=10,r_2=18)$. The two robots were usually assigned to different tasks to increase the expected pure reward, yet there are instances when they were assigned to the same task. The magenta curve shows the expected pure reward (up to the largest $t_f$ of the existing tasks), which would change as new tasks emerged and the robot agents were assigned to new tasks.
\section{Conclusion}
We propose the decentralized task and path planning (DTPP) framework that is capable of task allocation and high-level path planning for a multi-robot system in a fully decentralized manner. Each robot agent is modeled as a Mixed Observed Markov Decision Process (MOMDP) assuming the independent evolution of the robot states. The task allocation is solved by representing the total pure reward as a factor graph and using the max-sum algorithm to decentrally solve for the optimal solution. Potential conflicts between robot agents are resolved by a local forward dynamic programming scheme, which guarantees no collision between agents.

\balance
\renewcommand{\baselinestretch}{0.9878}
\bibliographystyle{myieeetran}
\bibliography{my_bib}
\appendix
\section{Optimal qualitative policy for MOMDP}\label{sec:quant}
Here we briefly describe how we approximate the optimal solution to the following cost optimal qualitative control problem
\begin{equation}\label{eq:minTimeQua}
\begin{aligned}
\underset{\pi}{\text{maximum}} ~~ & \mathbb{E}^\pi \Bigg[ \sum_{t = 0}^{t_f} -J(s_{t},a_{t},s_{t+1},o_{t+1}) \Bigg]   \\
 \text{subject to} ~~ ~& \pi \in \argmax_{\kappa} \mathbb{P}^\kappa[\bigvee_{t=0}^{t_f}s_t\in\G]. 
\end{aligned}
\end{equation}
Notice that the above qualitative constraint on the probability of satisfying the specifications can be rewritten as
\begin{equation*}
\begin{aligned}
    \mathbb{P}^\kappa[\bigvee_{t=0}^{t_f}s_t\in\G]&=\mathbb{P}^\kappa\big[ \exists k\in\{0,\ldots, t_f\}: s_k \in \G \big] \\
    &=\mathbb{E}^\kappa\Bigg[ \sum_{t=0}^{t_f} \Bigg( \prod_{\tau=0}^{t-1} \mathds{1}_{\mathcal{S} \setminus \G}(s_\tau) \Bigg) \mathds{1}_{\G}(s_t) \Bigg].
\end{aligned}
\end{equation*}
Furthermore, leveraging the result form~\cite[Lemma~4]{summers2010verification}, we have that the value function associated with the above reachability problem is given by the following recursion
\begin{equation}\label{eq:recursion}
\begin{aligned}
    V_\G^\kappa(t,s,\be) & = \mathds{1}_{\G}(s) + \mathds{1}_{\mathcal{Q}\setminus\G}(s) \mathbb{E}^\kappa[V_\G^\kappa(t+1,s',\be')] \\
    & = \begin{cases} 1 = \sum_{i = 1}^{|\mathcal{E}|}\be(i) & \mbox{ If } s\in \G \\
    \mathbb{E}^\kappa[V_\G^\kappa(t+1,s',\be')] & \mbox{ Else }
        \end{cases}
\end{aligned}
\end{equation}
with $V_\G^\kappa(t_f,s,\cdot)=1$ if $s\in \G$ and $V_\G^\kappa(t_f,s,\cdot)=0$ if $s\notin \G$.
Notice that $V_\G^\kappa(t_f,s,\cdot)$ is a linear function for all $s \in \mathcal{S}$ and, consequently, $V_\G^\kappa(t,s,\cdot): \Be \rightarrow \mathbb{R}$ is piecewise-affine by standard POMDP arguments~\cite[Theorem~7.4.1]{krishnamurthy2016partially}.

Finally, we have that as the value function~\eqref{eq:recursion} is piecewise-affine we can rewrite the quantitative problem~\eqref{eq:minTimeQua} as a constrained POMDP, which we approximated using modified version of the algorithm presented~\cite{isom2008piecewise}. We use another value function $V_J$ to keep track of the expected reward-to-go, which is minimized among all policies that maximize $V_\G$. In particular, compared to the algorithm presented in~\cite{isom2008piecewise}, we propagate only a single belief point per constraint. This strategy, while being sub-optimal, allows us to reduced the computational burden associated with the algorithm presented in~\cite{isom2008piecewise}.
\end{document}